\documentclass[sigconf]{acmart}
\usepackage{amsthm}
\usepackage{latexsym}
\usepackage{graphicx}
\usepackage{subfigure} 
\usepackage{url}
\usepackage{enumerate}
\usepackage{amsmath}
\usepackage{mathtools}
\usepackage{amssymb}
\usepackage{hyperref}

\DeclareMathOperator*{\argmax}{argmax}

\usepackage{tabularx, booktabs}
\newcolumntype{Y}{>{\centering\arraybackslash}X}
\newcolumntype{L}{>{\arraybackslash}X}
\newcolumntype{R}{>{\raggedleft\arraybackslash}X}

\usepackage{multirow}
\usepackage{xspace}

\usepackage{booktabs} 
\newcommand{\hide}[1]{} 

\newcommand{\ie}{\textit{i.e. }}
\newcommand{\eg}{\textit{e.g. }}

\newcommand{\nhnet}{\textsf{NHNet}}
\newcommand{\newshead}{\textsf{NewSHead}}
\newcommand{\Newshead}{\textsf{N{\small ew}SH{\small ead}}}
\newcommand{\newsheaddist}{\textsf{NewSHead\textsubscript{dist}}}

\definecolor{gred}{RGB}{219,68,55}
\definecolor{gblue}{RGB}{66,133,244}
\definecolor{gyellow}{RGB}{244,180,0}
\definecolor{ggreen}{RGB}{15,157,88}
\definecolor{ggrey}{RGB}{115,115,115}

\newcommand{\error}[1]{\textcolor{gred}{\textbf{#1}}} 
\newcommand{\fph}[1]{\textcolor{gblue}{\textbf{#1}}} 

\AtBeginDocument{%
  \providecommand\BibTeX{{%
    \normalfont B\kern-0.5em{\scshape i\kern-0.25em b}\kern-0.8em\TeX}}}

\setcopyright{none}
\renewcommand\footnotetextcopyrightpermission[1]{}
\copyrightyear{2020}
\acmYear{2020}
\setcopyright{rightsretained}

\acmConference{The Web Conference '20}{April 20 - 24, 2020}{Taipei}
\acmBooktitle{WWW '20, April 20 - 24, 2020, Taipei}

\newcommand{\HiddenVec}[2]{H_{#1}^{#2}}
\newcommand{\Article}{{\bf{a}}}
\newcommand{\Articles}{\mathcal{A}}
\newcommand{\Title}{\mathcal{T}}
\newcommand{\SeqTgt}{{\bf{y}}}
\newcommand{\TokenTgt}{y}

\newcommand{\eat}[1]{}

\renewcommand{\eg}{{\it e.g., }}
\renewcommand{\ie}{{\it i.e., }}

\begin{document}

\title{Generating Representative Headlines for News Stories}

\author{Xiaotao Gu$^{1 *}$, 
Yuning Mao$^1$, Jiawei Han$^1$, Jialu Liu$^2$, Hongkun Yu$^2$, You Wu$^2$, Cong Yu$^2$, }
\author{Daniel Finnie$^3$, Jiaqi Zhai$^3$, Nicholas Zukoski$^3$}
\affiliation{
  \institution{$^1$University of Illinois at Urbana-Champaign \quad
  $^2$Google Research \quad
  $^3$Google Inc.
  }
\institution{$^1$\{xiaotao2,yuningm2,hanj\}@illinois.edu
$^2$\{jialu,hongkuny,wuyou,congyu\}@google.com 
$^3$\{danfinnie,jiaqizhai,nzukoski\}@google.com 
}
  }
\thanks{\hspace{.3em} * Work done while interning at Google. Corresponding Author: Jialu Liu.} 
\renewcommand{\shortauthors}{Gu,~et.~al.}

\begin{abstract}
Millions of news articles are published online every day, which can be overwhelming for readers to follow. Grouping articles that are reporting the same event into \emph{news stories} is a common way of assisting readers in their news consumption.
However, it remains a challenging research problem to efficiently and effectively generate a representative \emph{headline} for each story.
Automatic summarization of a document set has been studied for decades, while few studies have focused on generating representative headlines for a set of articles.
Unlike summaries, which aim to capture most information with least redundancy, headlines aim to capture information jointly shared by the story articles in short length and exclude information specific to each individual article.

In this work, we study the problem of generating representative headlines for news stories.
We develop a distant supervision approach to train large-scale generation models without any human annotation. The proposed approach centers on two technical components. First, we propose a multi-level pre-training framework that incorporates massive unlabeled corpus with different quality-vs.-quantity balance at different levels. We show that models trained within the multi-level pre-training framework outperform those only trained with human-curated corpus. Second, we propose a novel self-voting-based article attention layer to extract salient information shared by multiple articles. We show that models that incorporate this attention layer are robust to potential noises in news stories and outperform existing baselines on both clean and noisy datasets.
We further enhance our model by incorporating human labels, and show that our distant supervision approach significantly reduces the demand on labeled data.
Finally, to serve the research community, we publish the first manually curated benchmark dataset on headline generation for news stories, \newshead{}, which contains $367$K stories (each with $3$-$5$ articles), $6.5$ times larger than the current largest multi-document summarization dataset.
\end{abstract}

\maketitle

\section{Introduction}
\label{sec:intro}

Today's news consumers are inundated with news content---over two million news articles and blog posts are published everyday\footnote{\href{http://www.marketingprofs.com/articles/2015/27698/2-million-blog-posts-are-written-every-day-heres-how-you-can-stand-out}{http://www.marketingprofs.com/articles/2015/27698/2-million-blog-posts-are-written-every-day-heres-how-you-can-stand-out}}.
As a result, services that organize news articles have become popular among online users. One approach is to categorize articles into pre-defined \emph{news topics}, each with a short category label, such as ``Technology,'' ``Entertainment'', and ``Sports''. While more organized, redundant content can still appear within each topic.
Another more effective approach is to further group articles within each topic based on \emph{news stories}. Here, each story consists of a cluster of articles reporting the same event. News stories make it more efficient to complete a reader's news consumption journey --- the reader can move from story to story and dive into each one as desired.
However, when we only present a list of article titles, readers can hardly get the gist of a story until they have read through several articles, as article titles are tailored to specific articles and do not provide an overview of the entire story.
Also, titles can be too long to scan through, especially on mobile devices.

\begin{figure}[t]
    \hspace{-0.3cm}\includegraphics[width=1.03\linewidth]{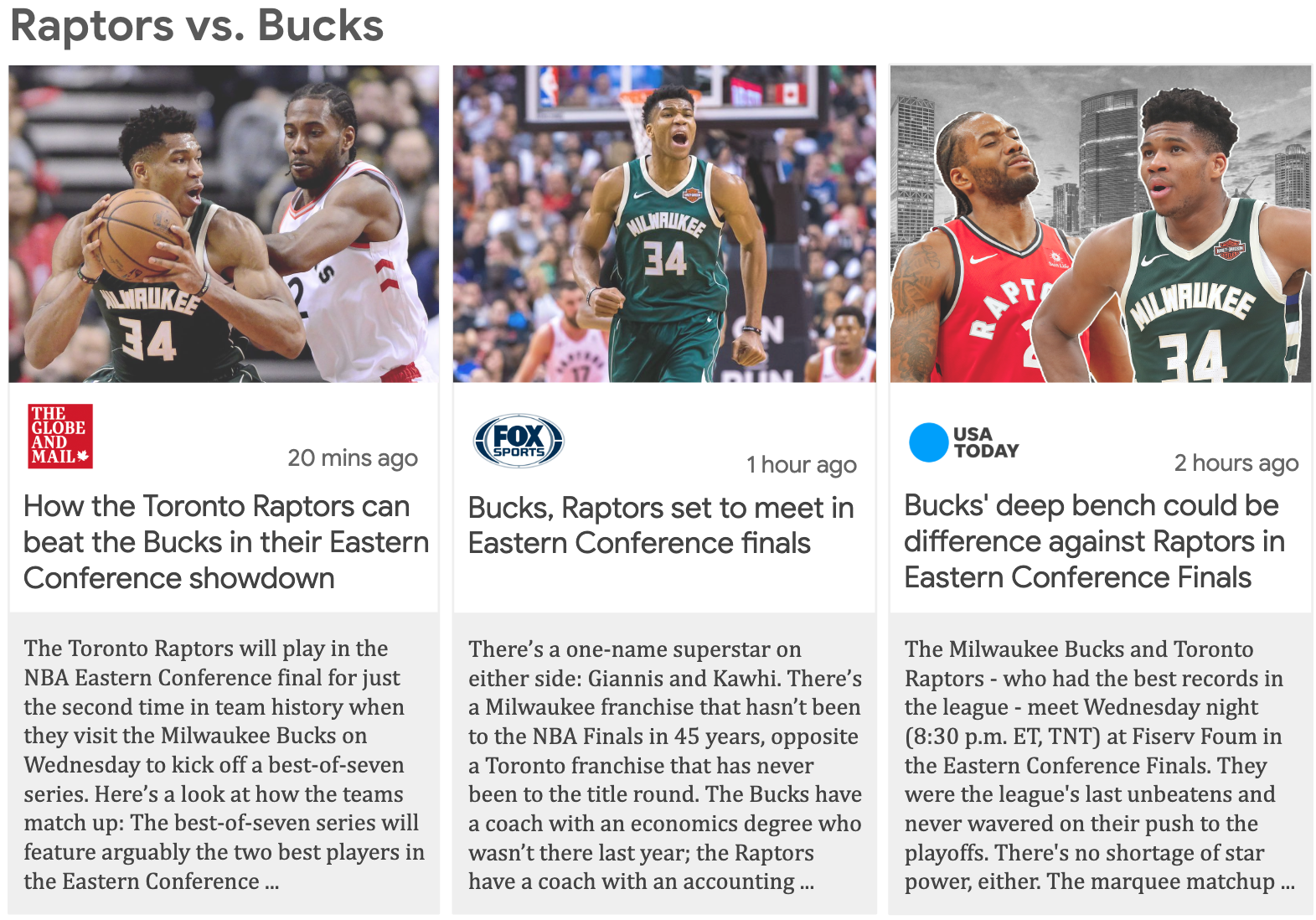}
    \vspace{-.3cm}
    \caption{An example of automatically generated story headline for articles about ``Raptors vs. Bucks''.}
    \label{fig:problem}
\end{figure}

To tackle this problem, we propose to summarize news stories with succinct and informative headlines. For example, ``Raptors vs. Bucks'' in Figure~\ref{fig:problem} headlines the cluster of news articles regarding the game between the two teams. Intuitively, headline is a useful complement to news clusters---users can quickly identify what stories they are planning to read in depth, without skimming a flotilla of unorganized news feeds. In practice, the value of headlines is also affirmed: Google News attaches a headline at the top of its story full coverage page\footnote{\href{https://www.blog.google/products/news/new-google-news-ai-meets-human-intelligence/}{\scriptsize https://www.blog.google/products/news/new-google-news-ai-meets-human-intelligence/}}.
However, it remains a challenging research problem to automatically generate story headlines.
First of all,
 selecting an existing article title may not be suitable, as article titles can be \emph{too long} (especially for readers on mobile devices) or \emph{unilateral} (incomplete perspective) to cover the general information of the story.
Next, relying on human editors to curate high-quality story headlines is inefficient, as it is not only expensive but also hard to scale due to the vast amount of emerging news stories and strict latency requirements.

To this end, we propose to study the problem of automatically generating representative headlines for news stories.
The main idea behind it, known as ``document summarization''\footnote{Article and document are used interchangeably throughout this paper. We use the former in the context of news and the latter for document summarization.}, has been studied for over six decades \cite{luhn1958automatic}.
The summarization task is to compress a single article into a concise summary while retaining its major information and reducing information redundancy in the generated summary \cite{filatova2004event,mihalcea2005language,woodsend2010automatic}.
As a special task of summarization, \textit{single-document} headline generation has also been thoroughly studied, whose generated summaries are not longer than one sentence \cite{banko2000headline,dorr2003hedge,zajic2002automatic}.
Recently, end-to-end neural generation models have brought encouraging performance for both abstractive summarization \cite{cheng2016neural,nallapati2016abstractive,narayan2018ranking,zhang-etal-2018-neural,see2017get} and single-document headline generation \cite{gavrilov2019self,hayashi2018headline,rush-etal-2015-neural,nallapati2016abstractive}.
Summarization of multiple documents has also gained much attention \cite{carbonell1998use,barzilay1999information,haghighi2009exploring,liu2018generating,fabbri-etal-2019-multi}, while headline generation for \textit{a set of documents} remains a challenging research problem.

The main challenge comes from the lack of high-quality training data.
Current state-of-the-art summarization models are dominated by neural encoder-decoder models \cite{you2019improving,zhang-etal-2018-neural,hayashi2018headline,gavrilov2019self}.
The high performance of these data-hungry models is supported by massive annotated training samples.
For single-document headline generation, such training data can be easily fetched from unlimited news articles with little cost: existing article-title pairs form perfect training samples.
Unfortunately, such a mapping is unavailable for multi-document settings.
Manually writing the summary or headline of a set of documents is also much more time consuming than that in the context of single-document summarization.
Hence, recent multi-document summarization (MDS) models either seek to adapt from single-document models \cite{lebanoff2018adapting,zhang2018adapting} or leverage external resources such as Wikipedia pages \cite{liu2018generating,liu-lapata-2019-hierarchical}.
\cite{fabbri-etal-2019-multi} recently provides a crowd-sourced dataset for multi-document summarization, but such resources remain absent for multi-document headline generation.

To facilitate standard study and evaluation, we publish the first dataset for multi-document headline generation.
The published dataset consists of 367K news stories with human-curated headlines, 6.5 times larger than the biggest public dataset for multi-document summarization~\cite{fabbri-etal-2019-multi}.
Large as it may seem, 367K news stories is still a drop in the ocean compared with the entire news corpus on the Web.
More importantly, manual curation is slow and expensive, and can hardly scale to web-scale applications with millions of emerging articles every day.
To this end, we propose to further leverage the unlabeled news corpus in two ways. 
Existing articles are first treated as a knowledge base and we automatically annotate unseen news stories by distant supervision (\ie with one of the article titles in the news story).
We then propose a multi-level pre-training framework, which initializes the model with a language model learned from the raw news corpus, and transfers knowledge from single-document article-title pairs.
The distant supervision framework enables us to generate another dataset for training without human effort, which is $6$ times larger than the aforementioned human-curated dataset.
We show that our model solely based on distant supervision can already outperform the same model trained on the human-curated dataset.
In addition, fine-tuning the distantly-trained model with a small number of human-labeled examples further boosts its performance (Section \ref{sec:finetune}).
In real-world applications, the process of grouping news stories, which is viewed as a prerequisite, is not always perfect.
To tackle this problem, we design a self-voting-based document-level attention model, which proves to be robust to the noisy articles in the news stories (Section \ref{sec:exp_noise}).
Improving the quality of clustering is out of the scope of this work, but remains an interesting future direction.

\smallskip
\noindent Our contributions are summarized as follows:
\begin{enumerate}
    \item We propose the task of headline generation for news stories, and publish the first large-scale human-curated dataset to serve the research community\footnote{\textbf{Data}: \url{https://github.com/google-research-datasets/NewSHead}}\footnote{\textbf{Code}: \url{https://github.com/tensorflow/models/tree/master/official/nlp/nhnet}};
    \item We propose a distant supervision approach with a multi-level pre-training framework to train large-scale generation models without any human annotation. 
    The framework can be further enhanced by incorporating human labels, but significantly reduces the demand of labels; 
    \item We develop a novel self-voting-based article attention module, which can effectively extract salient information jointly shared by different articles, and is robust to noises in the input news stories.
\end{enumerate}
\section{Problem Description}
Given a news story $\mathcal{A}$ as a cluster of news articles regarding the same event, where each article $\Article\in\mathcal{A}$ is composed of a token sequence $[a^1, a^2, \ldots]$,  we aim to generate a succinct and informative headline of the story, represented as another token sequence $\SeqTgt=[y^1, y^2, \ldots]$, such as ``Raptors'': $y^1$, ``vs.'': $y^2$ and ``Bucks'': $y^3$ in Figure~\ref{fig:problem} for a list of articles discussing about the series between the two teams.

Although the original title of each article can be a strong signal for headline generation, it is not included in our model input as (1) it increases the risk of generating clickbaity and biased headlines, and (2) high-quality titles can be missing in some scenarios (\eg user-generated content).
For these reasons, we only consider the main passage of each article as model input at this stage.

\section{The \Newshead{} Dataset}\label{sec:newshead}

\begin{table}[t]
    \centering
    \caption{Statistics comparison with MDS datasets.}
    \vspace{-.2cm}
    \label{tb:dataset}
    \scalebox{.99}{
    \renewcommand{\arraystretch}{0.6}
      \begin{tabular}{lcc}
        \toprule
        \textbf{Dataset} & \textbf{\# Instances} & \textbf{\begin{tabular}[x]{@{}c@{}} \# Characters{} in target\end{tabular}}\\
        \toprule
        \newshead{}  & 367,099 & 27.76\\
        \newsheaddist{} &  2,204,224&  38.37\\
        \midrule\midrule
        TAC2011 \cite{owczarzak2011overview} & 176 & 642.36 \\
        DUC03+04 \cite{paul2004introduction} & 320 & 675.43\\
        Multi-News \cite{fabbri-etal-2019-multi}   & 56,216 & 1298.62\\
        \bottomrule
    \end{tabular}%
    }
\end{table}

\begin{figure}
\centering     
\hspace{-0.1cm}\subfigure[Topic Distribution]{\label{fig:a}\includegraphics[width=.48\columnwidth]{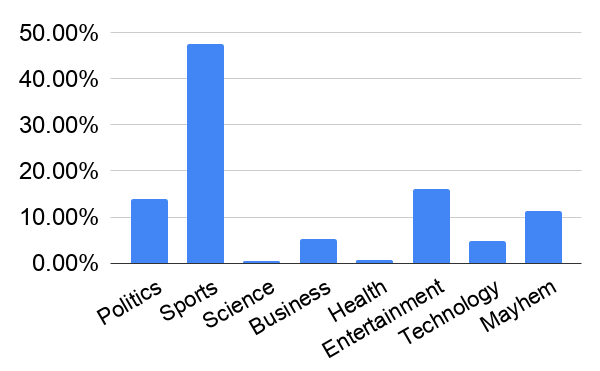}}\hspace{0.1cm}
\subfigure[Story Headline Length]{\label{fig:b}\includegraphics[width=.48\columnwidth]{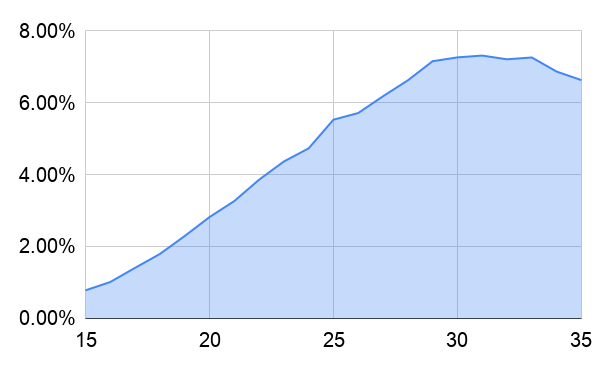}}
\hspace{-0.1cm}\subfigure[Article Title Length]{\label{fig:c}\includegraphics[width=.48\columnwidth]{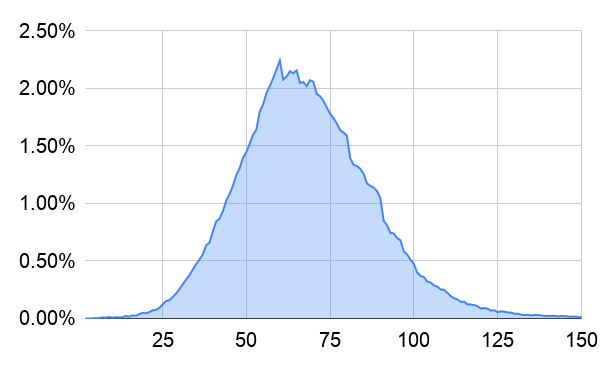}}\hspace{0.1cm}
\subfigure[Representative Title Length]{\label{fig:d}\includegraphics[width=.48\columnwidth]{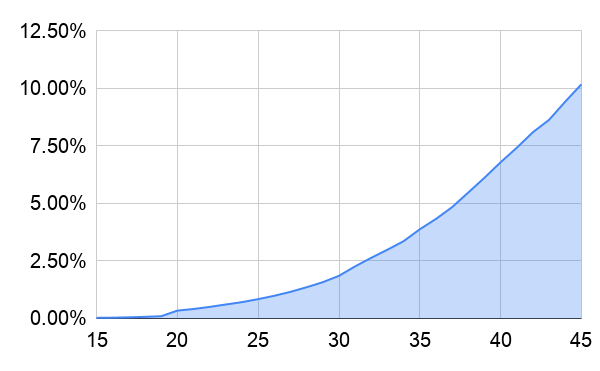}}
\caption{Visualization of data statistics: (a) topic distribution of articles in \newshead{}; (b) length distribution of manually curated story headlines (in character); (c) length distribution of the origianl article titles; (d) length distribution of selected representative titles in \newsheaddist{}.}
\end{figure}

To help future research and evaluation, we publish, to the best of our knowledge, the first expert-annotated dataset, \newshead{}, for the task of \textbf{New}s \textbf{S}tory \textbf{Head}line Generation.
The \newshead{} dataset is collected from news stories published between May 2018 and May 2019.
\newshead{} includes the following topics: Politics, Sports, Science, Business, Health, Entertainment, and Technology as depicted in Figure~\ref{fig:a}. A proprietary clustering algorithm iteratively
loads articles published in a recent time window and groups them based on content similarity. 
For each news story, curators from a crowd-sourcing platform are requested to provide a headline of \textit{up to 35 characters} to describe the major information covered by the articles in each story.
The curated headlines are then validated by other curators before they are included in the final dataset. Note that a story may contain hundreds of articles, and it is not realistic to ask curators to read through all the articles before curating a headline. Thus, only three to five representative articles that are close to the cluster centroid are picked to save human efforts.

Table~\ref{tb:dataset} shows the statistics of our dataset and the existing datasets for multi-document summarization (MDS).
In \newshead{}, each news story consists of 3-5 news articles.
This gives us 367K data instances, which is $6.5$ times larger than the biggest dataset for multi-document summarization \cite{fabbri-etal-2019-multi}.
We split the dataset by timestamps: the timestamps of all articles in the validation set are strictly greater than those in the training set. The same goes for the test set vs. validation set.
By avoiding overlapped time window, we can penalize overfitted models that memorize observed labels.
Overall, we generate 357K stories for training, 5K stories for validation, and 5K stories for testing, respectively.
As for the human-curated reference labels, as Table~\ref{tb:dataset} shows, the lengths of curated story headlines are much shorter than traditional summaries, and even shorter than article titles in our dataset depicted in Figure~\ref{fig:c}.
Figure~\ref{fig:problem} shows an example of a curated news story.
The story headline is much more concise than article titles in the cluster, and covers only general information shared by the articles.

Large as it may seem, the training data in \newshead{} remains a drop in the ocean compared with the entire news corpus, still leaving substantial room for performance improvement, under the assumption that modern models can achieve better performance with more data \cite{radfordimproving,liu2019roberta}.
Nevertheless, manual annotation is slow and expensive.
The amount of work and resource needed to create the \newshead{} dataset is already so much that scaling it up to just $500K$ instances seems to be cost-prohibitive. 
Facing this practical challenge, in the next section,
we will present a novel framework based on distant supervision to fetch additional training data without human annotation.

\section{Learning Framework}\label{sec:learning_framework}

\begin{figure}
    \includegraphics[width=0.8\columnwidth]{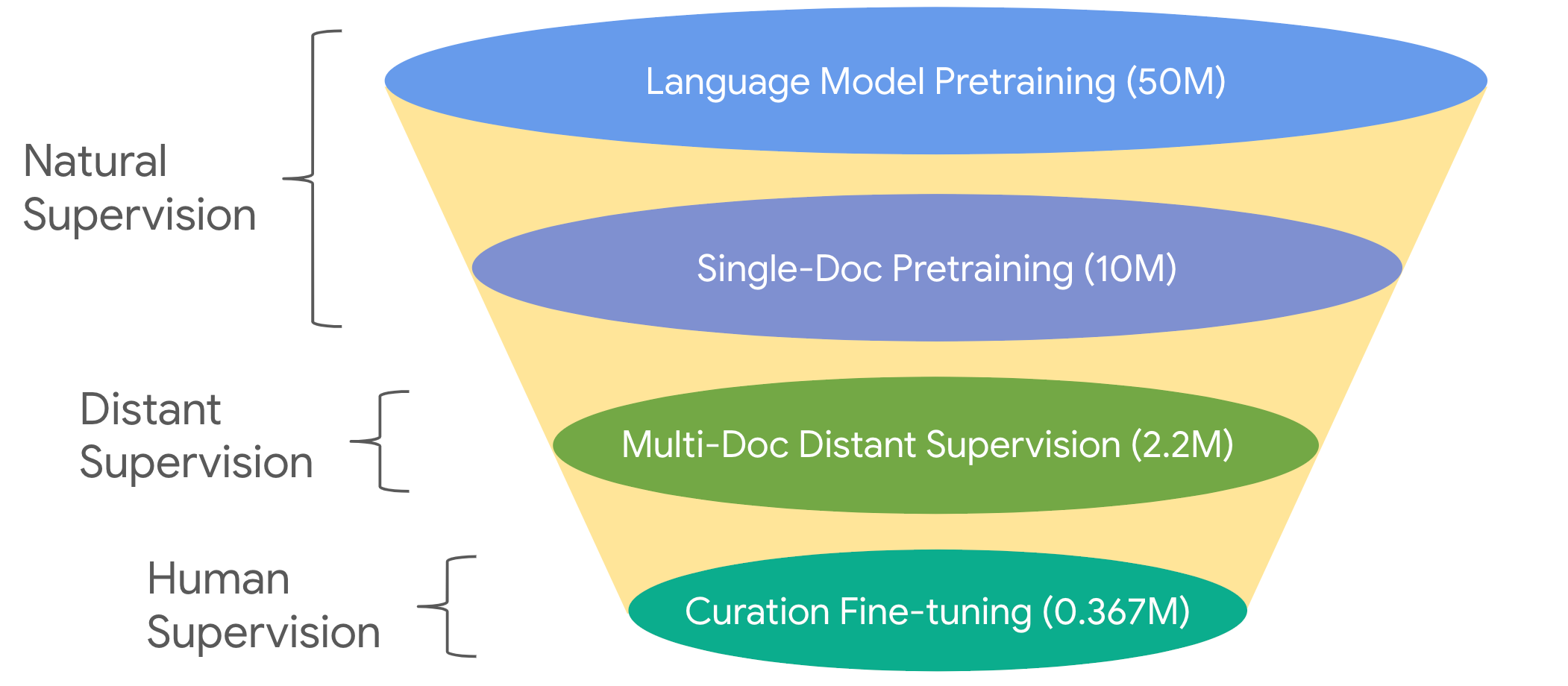}
    \caption{Multi-level supervision used in our framework.}
    \label{fig:pretrain}
\end{figure}
Learning an end-to-end generation model requires a large amount of annotated training data. 
Human annotation, however, is slow and expensive. It is thus hard for human annotation alone to provide sufficient data or scale to future scenarios.
In the following sections, we present a novel framework that leverages multiple levels of supervision signals to alleviate or even remove the dependency on human annotation.
As shown in Figure \ref{fig:pretrain}, we seek natural supervision signals in the existing news corpus to pre-train the representation learning and language generation module of our framework.
This process includes \emph{Language Model Pre-training} from massive text corpus, and transferring knowledge from article-title pairs (\emph{Single-Doc Pre-training}).
We then propose to generate heuristic training labels via \emph{Multi-Doc Distant Supervision}.
These supervision signals can be automatically fetched from the existing data with almost no cost. Later in Section~\ref{sec:experiment_result}, we show that models trained from these free signals can outperform those purely trained on the manually curated training set.

\subsection{Distant Supervision: \newsheaddist{}}
\label{sec:distant}
\begin{table}[t]
\footnotesize
    \caption{A comparison between the headline label generated by human curators in \newshead{} and the one generated by Distant Supervision in \newsheaddist{}.}
    \vspace{-.2cm}
    \label{tb:dataset_example}
    \scalebox{0.9}{
    \begin{tabular}{p{1\columnwidth}}
        \toprule
        \textbf{\newshead{}: {alabama senate abortion bill}}  \\
        \textbf{\newsheaddist{}: \fph{alabama senate approves near-total abortion ban}} \\
        \midrule
        
         \textbf{Title\textsubscript{1}}: \textbf{Alabama state Senate passes near total abortion ban in direct challenge to Roe v. Wade}\\
         The Senate rejected an attempt to add an exception for rape and incest, voting down the amendment 21-11. Alabama's Republican-dominated state Senate has passed a bill to ban nearly ... 
         
         \medskip
         \textbf{Title\textsubscript{2}}: \textbf{\fph{Alabama Senate approves near-total abortion ban}}\\
         The Alabama state Senate on Tuesday approved a bill essentially banning abortion in the state, a move specifically aimed at challenging more than 40 years...
         
         \medskip
         \textbf{Title\textsubscript{3}}: \textbf{Alabama Senate bans nearly all abortions, including rape cases}\\
         (Reuters) - Alabama's state Senate passed a bill on Tuesday to outlaw nearly all abortions, creating exceptions only to protect the mother's health, as part of a multistate effort to have the ...
         
         \medskip
         \textbf{Title\textsubscript{4}}: \textbf{Alabama state Democrat says near-total abortion bill passage `raped women last night'}\\
         (CNN) A Democratic state senator from Alabama equated a state bill that would ban abortion with no exceptions for rape and incest to the rape of Alabama women...

         \medskip
         \textbf{Title\textsubscript{5}}: \error{\textbf{25 men voted to advance most restrictive abortion ban in the country. The female governor signed it}}\\
         Washington (CNN) Only male Alabama senators voted Tuesday to pass the most restrictive abortion bill in the country, which criminalizes abortion in the state and bans the procedure in nearly all cases including rape and incest...\\
        \bottomrule
    \end{tabular}
    }
\end{table}
In this section, we show how to generate abundant training data from the existing corpus without human story curators.
A related technique, namely distant supervision, has proven to be effective in various information extraction tasks
(\eg Relation Extraction \cite{mintz2009distant} and Entity Typing \cite{ren2015clustype}).
The basic idea is to use existing \emph{task-agnostic labels} in a knowledge base (KB) to heuristically generate imperfect labels for specific tasks.
For example, in Entity Typing, one can match each entity mention (\eg \emph{Donald Trump}) in the sentence to an entry in the KB, and then label it with existing types of the entry (\eg \emph{Politician}, \emph{Human}, etc).
In this way, by leveraging existing labels and some heuristics, there is no need to spend extra time to curate labels for the specific task of interest.

Here, we view the news corpus as a KB, and treat existing article titles as candidate labels for news stories.
Note that not all article titles are suitable as story headlines, since, as we mentioned, some titles are too specific to cover the major information of the story.
Hence we need to automatically select high-quality story labels from many candidates without annotations from human experts.
Specifically, given a news corpus, we first group news articles into news stories.
This is an unsupervised process---the same as what we did for creating the \Newshead{} dataset.
For each story $\mathcal{A}$, we aim to get its heuristic headline $\hat{\SeqTgt}$ by selecting the most representative article title:
\begin{equation}
    \hat{\SeqTgt} = \argmax_{\Title_\Article : \Article \in \Articles} \frac{1}{|\Articles|} \sum_{\Article' \in \Articles \backslash \{\Article\}} f(\Title_\Article, \Article'),
\label{eq:distant}
\end{equation}
where $\Title_\Article$ stands for the title of article $\Article$ and $f(\Title, \Article)$ stands for the semantic matching score between any title $\Title$ and article $\Article$. Note that $\Article$ only contains tokens in the main passage.
In other words, the score of an article title is the average matching score between the title and other articles in the story.

The only problem now is to compute the matching score $f(\Title, \Article)$.
Instead of defining a heuristic score
(\eg lexical overlaps), we train a scorer with a binary classification task, where $f(\Title, \Article)$ is the probability that $\Title$ can be used to describe $\Article$.
Training data can be fetched by using existing article-title pairs as positive instances and sampling random pairs as negative instances.
We use a BERT-pre-trained Transformer model for this classification task with a cross-entropy loss.

We then follow Equation~\ref{eq:distant} to generate heuristic labels for unlabeled news stories.
It is likely that none of the article titles in the story is representative enough to be the story headline.
Hence, we only include positive labels\footnote{A label is positive if its average prediction score among all articles is above 0.5.} in the generated training data (leaving around 20\% of the stories).
The length distribution of the generated labels is shown in Figure \ref{fig:d}.
The average length is longer than human labels, but is in a reasonable area to generate enough training instances.
This way, we generate around $2.2$M labeled news stories without relying on human curators.
The new dataset, namely \newsheaddist{}, is $6$x larger than the annotated \newshead{} dataset.
Table \ref{tb:dataset_example} shows an example of heuristically generated labels in \newsheaddist{}.
Among all five candidate titles, the second title is ranked as the top choice since it well describes the general information of the story.
In comparison, the last title is not suitable as it is too specific and does not match some of the articles.

This way we can easily generate abundant high-quality training labels.
This generation process does not depend on human curators.
\newsheaddist{} is hence easy to scale as time goes by and the size of the news corpus gets larger.

\begin{figure*}[t]
    \includegraphics[width=\linewidth]{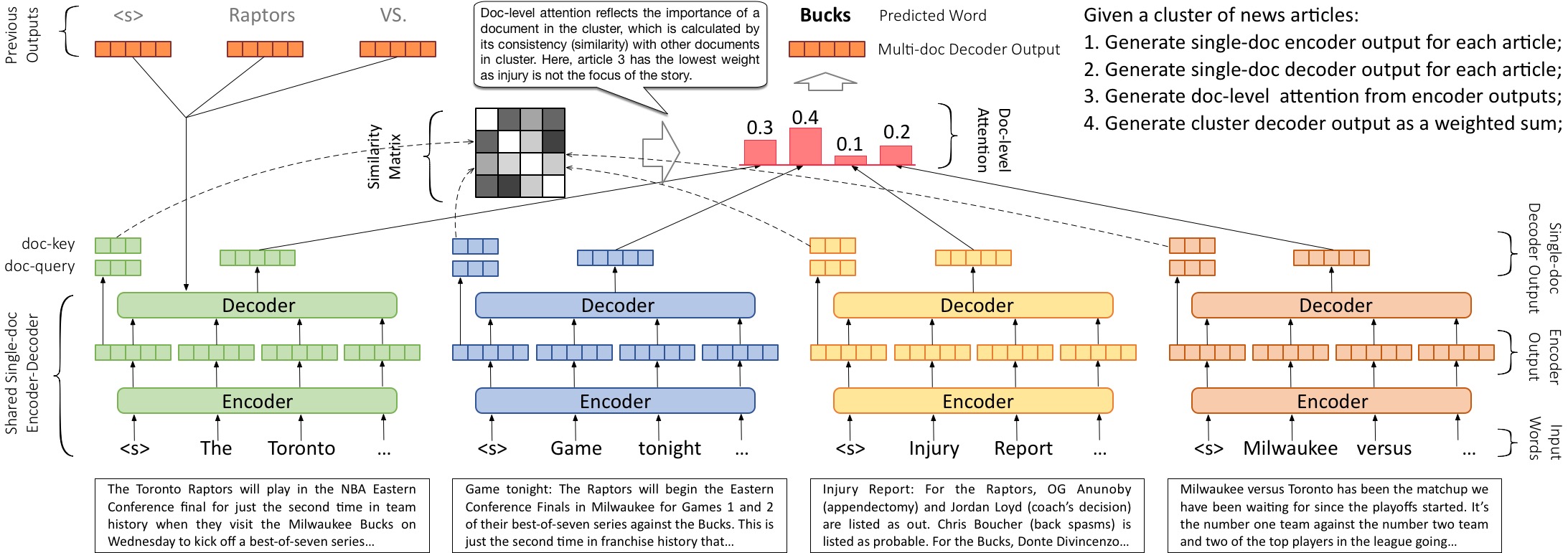}
    \vspace{-.4cm}
    \caption{The overall architecture of our model \nhnet. We use a shared encoder-decoder module to generate representation for each individual article, then integrate the results via an article-level self-voting attention layer (Section~\ref{sec:model}).}
    \label{fig:framework}
\end{figure*}

\subsection{Pre-training With Natural Supervision}
\label{sec:pretraining}
With the distantly supervised dataset comes a natural question: how far can we progress on the story headline generation task without human annotation?
To make the most of the massive unlabeled corpus, we apply pre-training techniques to enhance our model.
As an overview, our model includes an encoder-decoder module as its building block, and an article-level attention layer to integrate information from different articles in the story.
The detailed model architecture will be presented in Section~\ref{sec:model}.
At the pre-training stage, aside from \emph{Multi-Doc Distant Supervision}, we aim to initialize different modules with two kinds of \emph{Natural Supervision} signals from the existing news corpus as follows.

\smallskip
\noindent\textbf{Language Model Pre-training} transfers knowledge from the massive raw corpus in the news domain to enhance the representation learning module of our model. 
We followed the BERT pre-training paradigm \cite{devlin2019bert} to construct a dataset based on the main passages of over 50M news articles collected from the Web\footnote{One can use \href{http://commoncrawl.org}{CommonCrawl} to fetch web articles.}. This dataset consists of 1.3 billion sentences in total.
Both tasks of masked language model and next sentence prediction are included.
The learned parameters are used to initialize the encoder module and the word embedding layer. 
The decoder module can also be initialized with encoder parameters as suggested in literature \cite{rothe2019leveraging}. 
However, cross attention between the encoder and decoder modules remains uninitialized in this phase.

\smallskip
\noindent\textbf{Single-Doc Pre-training} leverages abundant existing article-title pairs to train the encoder-decoder module to enhance both representation learning and language generation.
In this step, we further adjust parameters for the encoder and decoder modules, together with cross attentions between them.
We clean the 50M raw news articles and leave only 10M high-quality article-title pairs for training.
For data cleaning, we first filter out article titles that are either too short (<15 characters) or too long (>65 characters).
We then use additional classifiers to remove article titles that are clickbaity, offensive or opinion-like. 
These additional classifiers are trained by binary labels collected from crowd-sourcing platforms.
Note that this filtering step is not mandatory in the framework.

\medskip
The model is further trained on distant supervision data with weights initialized from previous stages.
The previous two pre-training stages are used to initialize the encoder and decoder modules in the single document setting.
When it comes to multiple documents, where the model involves additional document attentions, we train such parameters together with previously mentioned model components.

Experiments show that models trained with above cost-free signals can even outperform models trained on manually curated training data.
By fine-tuning the model on human-curated labels, we can combine the two sources of supervision and further improve performance.

\section{The \nhnet{} Model}
\label{sec:model_overview}
In this section, we elaborate on the mathematical mechanism of our multi-doc news headline generation model, namely \nhnet{}.  We extend a standard Transformer-based encoder-decoder model to multi-doc setting and propose to use an article-level attention layer in order to capture information common to most (if not all) input articles, and provide robustness against potential outliers in the input due to clustering quality.  We analyze the model complexity compared to the standard Transformer model.

\subsection{Model Architecture}
\label{sec:model}

Figure \ref{fig:framework} illustrates the basic architecture of our generation model.
To take full advantage of massive unlabeled data, we start from a Transformer-based single-document encoder-decoder model as the building block.
The single-doc model generates decoding output for each article in the cluster individually.
To effectively extract common information from different articles, the model fuses decoding outputs from all articles via a self-voting-based article-level attention layer.
The framework proves to be not only easy to pre-train with distant supervision, but also robust to potential noises in the clustering process.

We start from a standard Transformer-based encoder-decoder model \cite{vaswani2017attention} as the building block.  In the single-doc setting, the sequence of tokens from an input article $\Article$ passes through a standard $L$-layer Transformer unit, with $h$ attention heads.  At decoding step $i$, the model takes in the full input sequence along with the output sequence produced up to step $i-1$ (\ie $[\TokenTgt^1,\TokenTgt^2,\dots,\TokenTgt^{i-1}]$), and yields a $d_H$-dimensional hidden vector (let it be denoted by $\HiddenVec{\Article}{i}$).  The end-to-end single-doc architecture would end up predicting the next token $\TokenTgt^i$ in the output sequence from $\HiddenVec{\Article}{i}$ as follows.
\begin{equation}
\label{eq:decode}
    P(\TokenTgt^{i}=y|[\TokenTgt^1,\TokenTgt^2,\dots,\TokenTgt^{i-1}]) = \frac{\exp(\HiddenVec{\Article}{i} \cdot M_y)}{\sum_{y' \in \mathcal{V}} \exp(\HiddenVec{\Article}{i} \cdot M_{y'})}
\end{equation}

Equation~\ref{eq:decode} above defines the probability of $\TokenTgt^{i}$ being $\TokenTgt$ given \linebreak $[\TokenTgt^1,\TokenTgt^2,\dots,\TokenTgt^{i-1}]$.  $\mathcal{V}$ is the entire vocabulary, $M_y$ is the column of a learnable embedding matrix $M \in \mathbb{R}^{d_H \times |\mathcal{V}|}$, in correspondence to token $y$.  Beam search is applied to find top-$k$ output sequences that maximize
\begin{equation*}
    \Pi_{i=1}^l P(\TokenTgt^i|[\TokenTgt^1,\TokenTgt^2,\dots,\TokenTgt^{i-1}]).
\end{equation*}

Extending to the multi-doc setting, we let each input article pass through the same Transformer unit and independently yield $\HiddenVec{\Article}{i}$, \ie apply the single-doc setting on all input articles up to the point of hidden vector representation.  Then we compute the vector representation of the input article group $\Articles$ as the weighted sum of $\{\HiddenVec{\Article}{i}\}$, i.e., $\HiddenVec{\Articles}{i} = \sum_{\Article\in\Articles} w_\Article\HiddenVec{\Article}{i}$.  The weights $\{w_\Article\}$ are determined by a similarity matrix learned via article-level attention (detailed next in Section~\ref{sec:article-attention}).  Finally, for predicting the next token $\TokenTgt^i$, we use $\HiddenVec{\Articles}{i}$ in place of $\HiddenVec{\Article}{i}$ in Equation~\ref{eq:decode}.

As we will show in Section~\ref{sec:article-attention}, the article-level attention introduces significantly fewer parameters to learn in addition to those in a standard Transformer model.

\subsection{Voting-based Article-level Attention}
\label{sec:article-attention}
The article-level attention layer is used to integrate information from all articles.
It assigns different attention weights to articles, which indicate the importance of each article.
To achieve this, previous works \cite{zhang2018adapting} use a learnable external query vector, denoted as the \emph{referee query vector} $Q_{r} \in \mathbb{R}^{d_H}$, to individually decide the weight of each article.
Specifically, the attention weight of article $\Article$ is computed as
\begin{equation}
    w_\Article = \frac{\exp(Q_r^T \cdot K_\Article)}{\sum_{\Article' \in \Articles} \exp(Q_r^T \cdot K_{\Article'})},
\end{equation}
where $K_\Article$ is the key vector of article $\Article \in \Articles$, which is usually linearly transformed from its encoded representations.
Such a design is intuitive, but ignores the interaction between articles.
For headline generation, our goal is to capture the common information shared by most articles in the story, where inter-article connections are important to consider.
More importantly, the clustered news story itself may not be perfectly clean: some articles may be loosely related, or even unrelated to other articles in the story.
In these scenarios, attention scores over articles should be determined by all articles together, instead of relying on an external referee.

\begin{figure}
    \includegraphics[width=0.8\columnwidth]{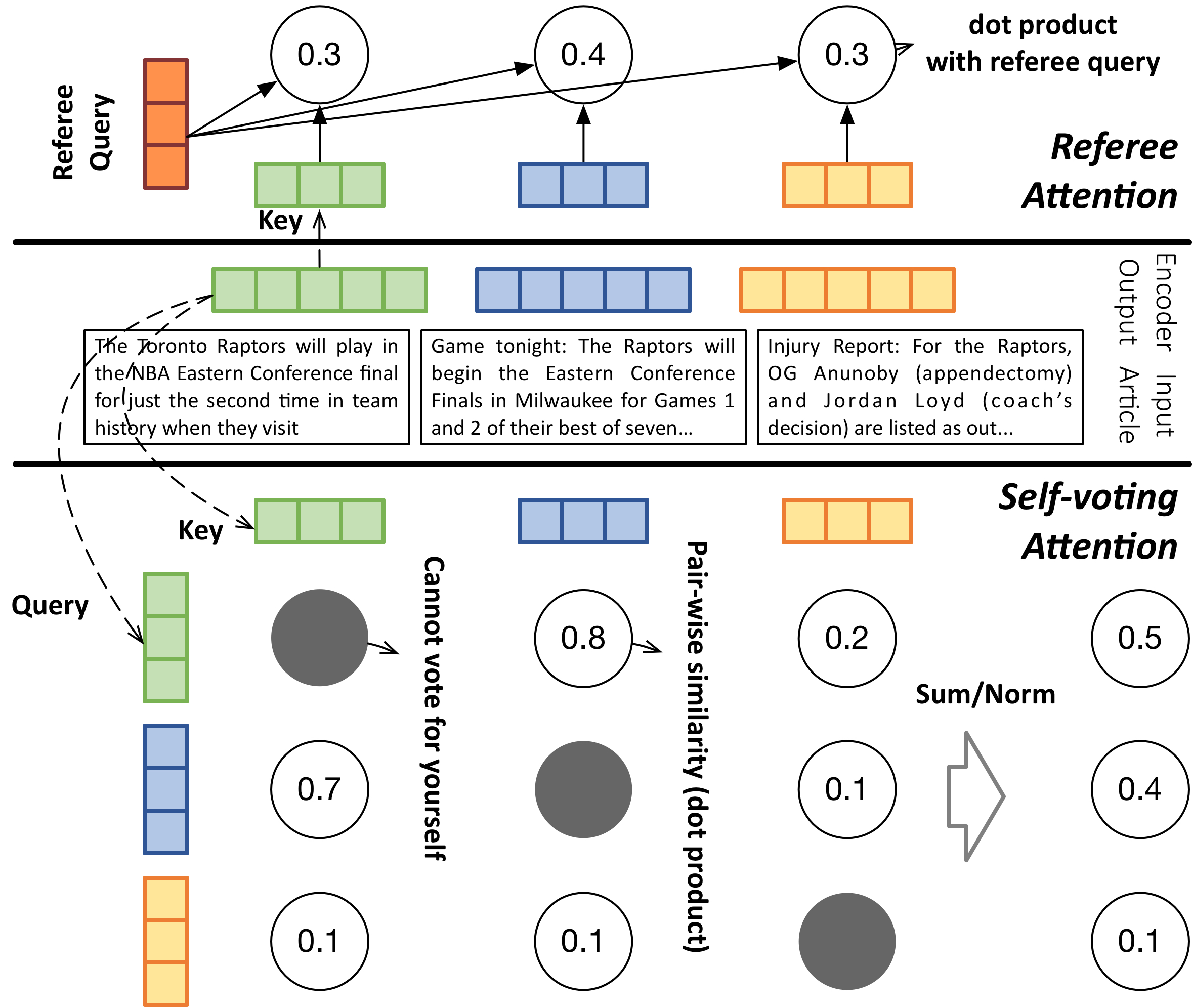}
    \vspace{-.2cm}
    \caption{Comparison between the referee attention and the proposed self-voting attention (Section~\ref{sec:article-attention}).}
    \label{fig:docatt}
\end{figure}

To this end, we design a simple yet effective \emph{self-voting-based article attention layer}.
The basic idea is to let each article vote for other articles in the story.
Articles with higher total votes from other articles shall have higher attention scores.
The advantages are two folds: common information shared by most articles will be echoed and amplified as it gets more votes, while irrelevant articles will be downplayed in this interactive process to reduce harmful interference on the final output.
Specifically, given the representation vector $H_\Article$ of article $\Article$, we calculate its query vector and key vector as $Q_\Article = W_Q \cdot H_\Article$ and $K_\Article = W_K \cdot H_\Article$, respectively, where $W_Q$ and $W_K$ are learnable matrices shared by all articles.
The attention score of $\Article$ is then computed as
\begin{equation}
    w_\Article = \frac{\exp(\sum_{\Article' \in \Articles\backslash \{\Article\}} Q_{\Article'}^T \cdot K_\Article)}{\sum_{\Article'' \in \Articles}\exp(\sum_{\Article' \in \Articles\backslash \{\Article''\}} Q_{\Article'}^T \cdot K_{\Article''})},
\end{equation}
where $\exp(Q_{\Article'}^T \cdot K_\Article)$ represents the vote of $\Article'$ on $\Article$.
Figure \ref{fig:docatt} illustrates the similarities and differences between the referee attention and self-voting attention.
Among the three example articles, the first two are introducing the game between the two teams, while the third one is focusing on injury information, which is too specific to be included in the headline.
Through the self-voting process, the article group finds that the third article is more distant from the central topic, and hence downplays its weight.
As a result, the generated headline focuses more on the game between the Raptors and Bucks instead of injury information.
The referee attention can hardly achieve the same goal as it ignores other articles when assigning the weight to each individual article.
As one can expect, the self-voting attention module is also more robust to potential noises in the cluster.
In the sanity checking experiments, the attention module usually gives weights close to zero for intentionally added noisy articles, demonstrating its capability in detecting off-topic articles, whereas the referee attention can hardly sense the differences.

\paragraph{Model complexity.}  The standard Transformer model consists of $O(L\cdot d_H\cdot h\cdot d_v)$ parameters, where $d_v$ is the dimension of projected value vector space, and $L$, $d_H$, $h$ are as defined in Section~\ref{sec:model}.  
Adding article-level attention introduces an additional $O(d_H\cdot h\cdot d_v)$ parameters---$1/L$ of what is already required by Transformer.

\section{Experimental Settings}

\begin{table*}[t]
    \centering
    \caption{Performance comparison of different methods. H stands for training with human-curated labels. L/S/D denotes Language model pre-training, Single-document pre-training and Distant supervision pre-training respectively.}
    \vspace{-.2cm}
    \label{tb:performance}
    \scalebox{0.93}{
    \renewcommand{\arraystretch}{0.8}
      \begin{tabular}{lccccccccccc}
        \toprule
        \textbf{Method} & R1-P & R1-R & \textbf{R1-F} & R2-P & R2-R & \textbf{R2-F} & RL-P & RL-R & \textbf{RL-F} & \textbf{Len-C} & \textbf{Len-W}\\
        \toprule
        \textbf{LCS} & 0.437 & 0.646 & 0.486 & 0.241 & 0.385 & 0.272 & 0.413 & 0.620 & 0.462 & 0.796 & 0.788\\
        \textbf{RepTitles} & 0.664 & 0.364 & 0.458 & 0.340 & 0.172 & 0.220 & 0.598 & 0.328 & 0.413 & 1.984 & 1.933\\
        \midrule
        \textbf{WikiSum} (H) & 0.561 & 0.591 & 0.567 & 0.312 & 0.326 & 0.313 & 0.509 & 0.535 & 0.514 & 0.958 & 0.966\\
        \textbf{SinABS} (H+LS) & 0.744 & 0.748 & 0.738 & 0.499 & 0.501 & 0.493 & 0.680 & 0.682 & 0.674 & 1.021 & 1.018\\
        \textbf{Concat} (H+LSD) & 0.752 & 0.756 & 0.746 & 0.510 & 0.510 & 0.503 & 0.689 & 0.694 & 0.685 & 1.017 & 1.013\\
        \textbf{SinABS} (H+LSD) & 0.758 & 0.769 & 0.755 & 0.522 & 0.530 & 0.518 & 0.695 & 0.704 & 0.692 & 1.004 & 1.006\\
        \midrule
        \textbf{\nhnet{}} (H) & 0.596 & 0.621 & 0.600 & 0.327 & 0.338 & 0.327 & 0.539 & 0.560 & 0.542 & 0.983 & 0.980\\
        \textbf{\nhnet{}} (LSD) & 0.726 & 0.590 & 0.639 & 0.440 & 0.354 & 0.382 & 0.667 & 0.542 & 0.588 & 1.327 & 1.286\\
        \midrule
        \textbf{\nhnet{}} (H+L) & 0.716 & 0.728 & 0.714 & 0.466 & 0.471 & 0.462 & 0.657 & 0.668 & 0.656 & 1.003 & 1.000\\
        \textbf{\nhnet{}} (H+LS) & 0.751 & 0.776 & 0.755 & 0.514 & 0.530 & 0.514 & 0.688 & 0.710 & 0.691 & 0.992 & 0.988\\
        \textbf{\nhnet{}} (H+LSD) & \textbf{0.762} & \textbf{0.779} & \textbf{0.762} & \textbf{0.531} & \textbf{0.542} & \textbf{0.529} & \textbf{0.703} & \textbf{0.718} & \textbf{0.703} & 1.003 & 0.997\\
        \bottomrule        
    \end{tabular}%
    }
\end{table*}
For standard evaluation, we compare all methods on the test set of \newshead, and tune parameters on the validation set.

As headline generation for news stories is a new task, instead of a ``state-of-the-art'' race between various models, we are more curious about the following questions regarding realistic applications:

\begin{enumerate}[(1)]
    \item \emph{How far} can we go without any human annotation but using existing natural supervision signals only?
    \item \emph{Can we} further boost the performance of distantly supervised models by incorporating human labels? 
    \item \emph{How} does the performance change with more human labels?
    \item \emph{How robust} are these methods to potential noises in news stories, as the story clustering process can be imperfect?
\end{enumerate}

\subsection{Baseline Methods}
Although no previous methods follow exactly the same setting of our task, variants of methods for multi-document summarization (MDS) can serve as our baseline methods with minor adjustments.
Specifically, we consider two families of models.

\smallskip
\noindent\textbf{Extractive Methods.} 
Extractive MDS models cannot be directly applied as they generate the summary by selecting sentences from the document set, while our expected output is a concise headline.
Extracting suitable words from article bodies is also challenging.
Here we consider two competitive baseline methods that (cheatingly) extract information from article titles:
\begin{itemize}
    \item \textbf{\textsf{LCS}} extracts the \textbf{l}ongest \textbf{c}ommon (word) \textbf{s}equence of article titles in the story. In case that no common word sequence exists, we relax the constraint of finding a common sequence shared by at least two articles.
    \item \textbf{\textsf{RepTitles}} uses the title scorer that we introduced in Section~\ref{sec:distant} for distant supervision, to select the most representative article title in the story as the predicted headline.
\end{itemize}
Note that the article titles are unavailable for abstractive models including our model.

\smallskip
\noindent\textbf{Abstractive Methods.}
Abstractive models take article bodies as input and generate the story headline in an end-to-end manner.
Since our full model makes use of different kinds of additional natural supervision (\eg unlabeled corpus and article-title pairs), it would be unfair to compare with models that are not designed to use such signals.
For illustration, we compare with both such traditional models (\eg WikiSum \cite{liu2018generating}) and models designed to leverage additional supervision for pre-training (SinABS \cite{zhang2018adapting}).
Additional enhancements are applied to baseline models to make them even stronger and more comparable to our model.
All methods are pre-trained with the same resources.

\begin{itemize}
    \item \textbf{\textsf{WikiSum}} \cite{liu2018generating}, as a representation of supervised abstractive models, generates summaries from the concatenation of the ordered paragraph list in the original articles.
    It proposes a decoder-only module with memory-compressed attention for the abstractive stage.
    \item \textbf{\textsf{Concat}} first concatenates body texts in all articles of a story, and then uses the single-document encoder-decoder building block of our model to generate the headline. This way, every attention layer in Transformer is able to access tokens in the entire story. To avoid losing word position information after concatenation, the positional encoding is reset at the first token for each article, so the model can still identify important leading sentences of each article.
    \item \textbf{\textsf{SinABS}} \cite{zhang2018adapting} is a recently proposed model which adapts and outperforms the state-of-the-art MDS models by pre-training on single-document summarization tasks. It uses a referee attention module to integrate encoding outputs from different articles as the representation for the article set. For fair comparison, we replace the original LSTM encoder with the transformer architecture, with the same parameter size as our model. 
    \item \textbf{\textsf{SinABS (enhanced)}}: The original model only makes use of knowledge from single articles. We further enhance it with distant supervision (H+LSD in Table \ref{tb:performance}).
    \item \textbf{\textsf{\nhnet{}}} is the model we proposed in Section~\ref{sec:model_overview}.
\end{itemize}

\noindent We test these baseline methods with different training settings, and report detailed performance comparison and analysis in Section \ref{sec:results}.

\subsection{Datasets}
As introduced in the multi-level training framework in Sections~\ref{sec:newshead} and \ref{sec:learning_framework}, the datasets used in this work include
\begin{enumerate}
    \item Language model Pre-training (\textbf{L}) with 50M articles;
    \item Single-Doc Pre-training (\textbf{S}) with 10M articles;
    \item Distant Supervision \newsheaddist{} (\textbf{D}) with 2.2M stories;
    \item Human annotations from \Newshead{} (\textbf{H}) with 357k stories.
\end{enumerate}
The entire \Newshead{} contains 367k instances, among which we use 5k for validation and 5k for final testing.

The labels for training and testing in this work are uncased, since both human-labeled and distantly supervised headline labels may contain various case formats, which can influence the learning and evaluation process.
In application, the task of recovering case information from generated uncased headlines, which is also called truecasing in natural language processing, is treated as a separate task. Meanwhile, we discover that simple majority voting from cased frequent n-grams in the story content is an accurate solution.

\subsection{Reproduction Details}
We use the WordPiece tool \cite{wu2016google} for tokenization. The vocabulary size is set to 50k and is derived from uncased news corpus.
Each sentence in the news article is tokenized into subwords, which significantly alleviates out-of-vocabulary (OOV) issues.
We also tried to incorporate copy mechanism \cite{gu2016incorporating}, another popular choice to reduce OOV, but did not see significant improvement. For better efficiency and less memory consumption, we use up to 200 WordPiece tokens per article as input.

As for the Transformer model, we adopt a standard ($L$=)12-layer architecture, with ($h$=)16 heads, hidden states of ($d_H$=)768 dimensions, and projected value space of ($d_v$=)48 dimensions.
For training, we use the Adam \cite{kingma2014adam} optimizer with a 0.05 learning rate and a batch size of 1024.
For every model we use the same early stopping strategy to alleviate overfitting: we first let the model train for 10k steps, after that we stop the training process once the model cannot steadily achieve higher performance on the validation dataset.

We implement the model in Tensorflow and train the model on cloud TPUs. Our code will be released together with the \Newshead{} dataset.

\subsection{Evaluation Metrics}

For evaluation we use the open source scoring tool \footnote{https://github.com/google-research/google-research/blob/master/rouge/scoring.py} that aggregates scores from a bootstrapping process. We report {\it{average}} results on the 5k \newshead{} test set.
We use the following metrics to evaluate the generated headlines:

\noindent \textbf{ROUGE} measures n-gram overlap between predicted headlines and gold labels.
Here we report the R-1, R-2, R-L scores in terms of (p)recision ($\frac{\# overlap}{\# predicted}$), (r)ecall ($\frac{\# overlap}{\# gold}$) and F1 ($\frac{2 \cdot p \cdot r}{p + r}$).

\smallskip
\noindent\textbf{Relative Length} measures the ratio between the length of predicted headlines and the gold labels ($\frac{LenPredict}{LenGold}$).
Here we report the ratio in terms of both words (\emph{\small Len-W}) and characters (\emph{\small Len-C}).
The closer the relative length is to $1.0$, the more likely it is that the generated headline has a similar length to the gold headline, whose distribution is shown in Figure~\ref{fig:b}.

\section{Experiment Results}\label{sec:experiment_result}
\begin{table*}[t]
    \centering
    \caption{Robustness Comparison of Different Attention Designs on the Noisy Dataset.}
    \vspace{-.3cm}
    \label{tb:ablation}
    \scalebox{0.95}{
    \renewcommand{\arraystretch}{0.6}
      \begin{tabular}{lccccccccccc}
        \toprule
        \textbf{Method} & R1-P & R1-R & \textbf{R1-F} & R2-P & R2-R & \textbf{R2-F} & RL-P & RL-R & \textbf{RL-F} & \textbf{Len-C} & \textbf{Len-W}\\
        \toprule
        \textbf{Uniform} (H) & 0.544 & 0.560 & 0.543 & 0.284 & 0.290 & 0.281 & 0.492 & 0.505 & 0.491 & 0.994 & 1.003\\
        \textbf{Uniform} (H+L) & 0.680 & 0.701 & 0.683 & 0.428 & 0.443 & 0.429 & 0.626 & 0.646 & 0.629 & 0.986 & 0.990\\
        \textbf{Uniform} (H+LS) & 0.736 & 0.747 & 0.733 & 0.494 & 0.500 & 0.490 & 0.672 & 0.681 & 0.669 & 1.011 & 1.011\\
        \textbf{Uniform} (H+LSD) & 0.742 & 0.759 & 0.743 & 0.507 & 0.519 & 0.506 & 0.682 & \textbf{0.697} & 0.682 & 0.999 & 0.998\\
        \midrule
        \textbf{Referee} (H) & 0.299 & 0.308 & 0.297 & 0.118 & 0.120 & 0.115 & 0.272 & 0.281 & 0.270 & 0.995 & 1.085 \\
        \textbf{Referee} (H+L) & 0.695 & 0.673 & 0.675 & 0.434 & 0.418 & 0.418 & 0.638 & 0.617 & 0.620 & 0.993 & 0.989 \\
        \textbf{Referee} (H+LS) & 0.737 & 0.719 & 0.719 & 0.484 & 0.472 & 0.470 & 0.672 & 0.655 & 0.655 & 0.999 & 0.996 \\
        \textbf{Referee} (H+LSD) & 0.739 & 0.750 & 0.737 & 0.496 & 0.503 & 0.492 & 0.675 & 0.684 & 0.673 & 1.009 & 1.010 \\
        \midrule
        \textbf{\nhnet{}} (H) & 0.565 & 0.594 & 0.571 & 0.307 & 0.320 & 0.307 & 0.512 & 0.538 & 0.517 & 0.972 & 0.972\\
        \textbf{\nhnet{}} (H+L)   & 0.692 & 0.707 & 0.692 & 0.434 & 0.444 & 0.432 & 0.631 & 0.645 & 0.631 & 0.996 & 1.000\\
        \textbf{\nhnet{}} (H+LS) & 0.745 & 0.756 & 0.743 & 0.506 & 0.512 & 0.502 & 0.684 & 0.693 & 0.681 & 1.010 & 1.007\\
        \textbf{\nhnet{}} (H+LSD) & \textbf{0.758} & \textbf{0.761} & \textbf{0.752} & \textbf{0.521} & \textbf{0.521} & \textbf{0.514} & \textbf{0.695} & \textbf{0.697} & \textbf{0.689} & 1.018 & 1.017\\
        \bottomrule
    \end{tabular}%
    }
\end{table*}

In the following, we answer the questions raised in the beginning of the former section.
\label{sec:results}
\subsection{Performance Comparison}
Table~\ref{tb:performance} shows the performance of compared methods trained with different combinations of datasets.
Generally, abstractive methods outperform extractive methods, even though the latter have access to already summarized information from existing article titles.
Among abstractive methods, the carefully designed concatenation model can achieve comparable performance with the existing state-of-the-art.
When enhanced with the distantly supervised training data (H+LSD), the SinABS model can have stronger performance.
Our model, when trained with the same resources, consistently outperforms baseline methods.

\subsection{How effective is distant supervision?}
To investigate the necessity of manual curation for this task, we compare the fully supervised model with human annotations only and the distantly supervised model with all natural supervision signals (\ie Language Model Pre-training and Single-Document Pre-training).
To our surprise, the distantly supervised model outperforms the fully supervised model by a significant margin, even though the distantly supervised training labels have different styles and lengths compared to those in the final test set.
The result is encouraging by revealing an effort-light approach for this task.
Instead of relying on human experts to curate story headlines for over a year, we can automatically mine high-quality headlines and natural supervision signals from existing news data for training in one day.
This observation can serve as a solid foundation for the future development of large-scale production models.
Fine-tuning the learned model with human labels can further boost performance.
As an ablation study, we investigate the model fine-tuned with full human annotation under different pre-training settings.
Starting from the fully supervised model with only human annotations (H), every pre-training process (+L+S+D) brings considerable improvements as we expect.

\subsection{How many manual labels do we need?}
\label{sec:finetune}
\begin{figure}
    \includegraphics[width=0.68\columnwidth]{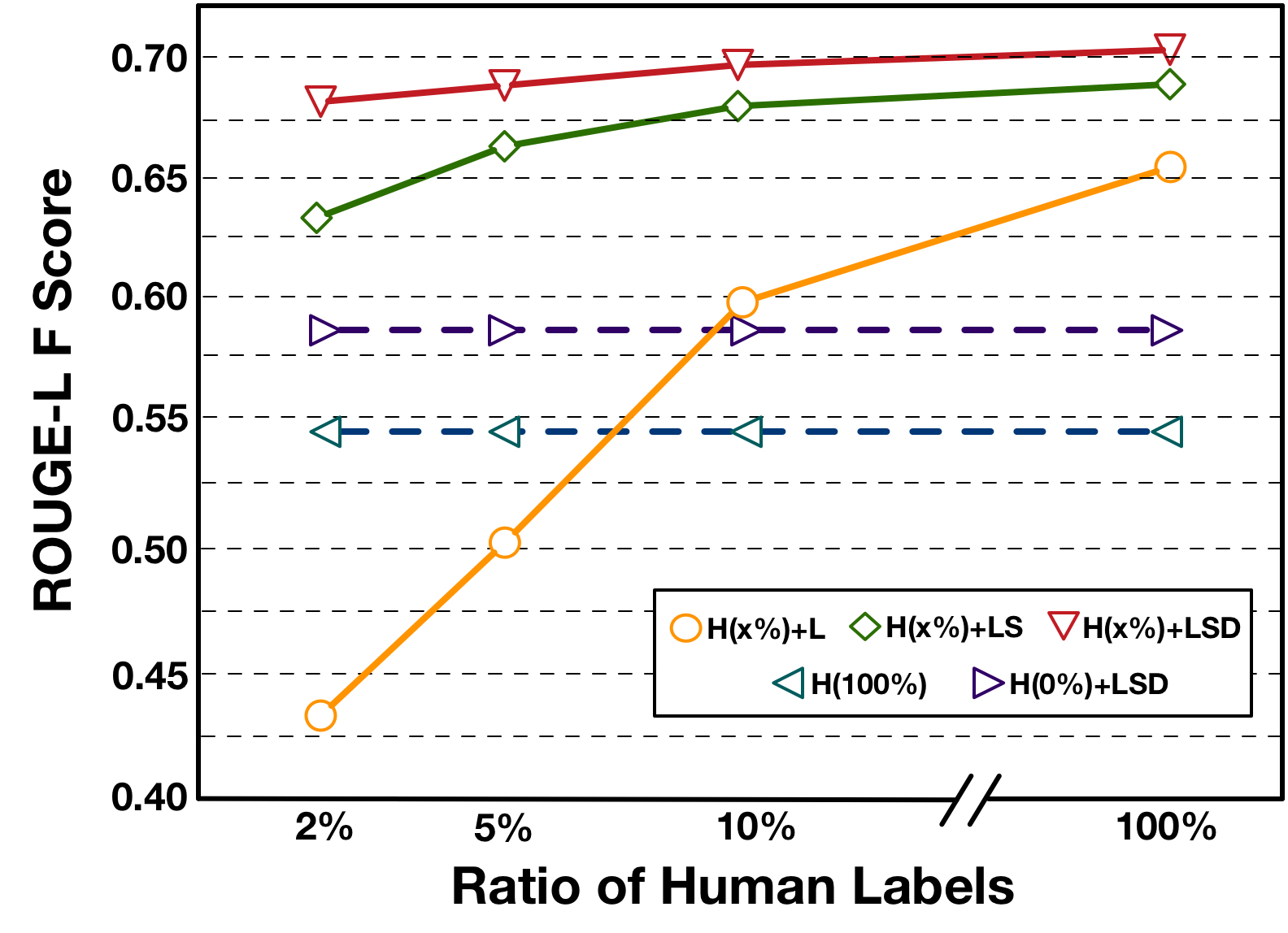}
    \vspace{-.2cm}
    \caption{An illustration of how performance changes with more manual labels involved for fine-tuning. X-axis stands for the ratio of training data used. Y-axis stands for the RL-F score on the test set.}
    \label{fig:finetune}
    \vspace{-.3cm}
\end{figure}
Our model achieves the best performance when combining human annotation with all kinds of distant and natural supervision (H($100\%$)+LSD).
Since the distantly supervised model may still need some manual labels to adjust the style and length of its generated headlines, it is hence meaningful to investigate the trade-off between the number of manual labels and the final performance to reasonably save human efforts.

Figure~\ref{fig:finetune} shows how test performance changes as we use different ratios of manual labels to fine-tune the distantly supervised model.
Generally, more manual labels lead to better test performance, as one can expect.
However, different models have various demand for the human labels to achieve the same performance.
As shown, when only $2\%$ of human labels are available, the supervised model, even with Language Model Pre-training (H($2\%$)+L), achieves significantly worse performance than the model trained with $100\%$ human labels.
Our model with distant supervision (H($2\%$)+LSD), on the contrary, outperforms the fully supervised model by a large margin.
More labels beyond this amount only bring slight improvements.
This further validates our idea: through distant supervision, we are able to learn high-quality models with very few manual labels.

\begin{table*}[t]
    \footnotesize
        \caption{Case studies on the effectiveness of distant supervision and human supervision in the clean dataset. At most three articles of each story are shown for the interest of space.}
        \vspace{-.17cm}
        \label{tb:case_clean}
        \scalebox{0.9}{
        \begin{tabular}{p{1.03\columnwidth} p{1.03\columnwidth}}
            \toprule
            \textbf{Gold Label}: {brazil's copa america squad} & \textbf{Gold Label}: {porter jr. to play in summer league}\\
            \textbf{Distant Supervision}: brazil announce squad for copa america & \textbf{Distant Supervision}: michael porter jr. will play in summer league \\
            \textbf{Distant Supervision+Human Supervision}: brazil copa america squad & \textbf{Distant Supervision+Human Supervision}: porter jr to play in summer league   \\
            \midrule
            
            \textbf{Article\textsubscript{1}}: \textbf{Neymar Leads Brazil's Copa America Squad; Ajax's David Neres Included}
            \newline               {RIO DE JANEIRO (AP) - Neymar and David Neres are in for the \fph{Copa America}, Lucas Moura and Vinicius Jr. are out. Brazil coach Tite named his squad Friday...}
            & \textbf{Article\textsubscript{1}}: \textbf{Michael Porter Jr. is pain-free and expects to play in Summer League}
            \newline
             {Early Monday morning, less than 24 hours after the Denver Nuggets season ended in heartbreaking fashion, the 14th overall pick of the 2018 draft, \error{Michael} \fph{Porter Jr.}, was made available to the media during exit interviews... }\\
            \smallskip
             \textbf{Article\textsubscript{2}}: \textbf{Vinicius Jr. not in Brazil's Copa America squad}
            \newline {\fph{Brazil} coach Tite has included Neymar in his 23-man \fph{squad} \error{for} the \fph{Copa America} his country will host but has yet to decide if the Paris-Saint Germain forward will retain the captaincy -- while Real Madrid's teenage forward Vinicius Junior did not make the cut...} 
             & \smallskip \textbf{Article\textsubscript{2}}: \textbf{Michael Porter Jr. Fully Cleared, Will Play In Summer League} \newline {\error{Michael} \fph{Porter Jr.} has been fully cleared for all basketball related activities and expects \fph{to play in Summer League}. "I can't wait for that," said \fph{Porter}. \fph{Porter} said he also could have played towards the end of the 18-19 season. \fph{Porter} was the 14th overall pick...} \\
            \smallskip
             \textbf{Article\textsubscript{3}}: \textbf{Brazil Announce Their Copa America Squad, Some Big Names Missing From The Team}
             \newline                 {\fph{Brazil} have \error{announced} their \fph{squad} ahead of the upcoming \fph{Copa America}. Head coach Tite will be hoping that his side are able to capture the trophy on home soil and he has a stacked \fph{squad} of players to use for the competition...} 
             & \smallskip \textbf{Article\textsubscript{3}}: \textbf{Denver Nuggets set to add Michael Porter Jr. to the mix next season}
            \newline {Even with an earlier playoff exit than they would have liked, after earning the No. 2 seed in the West, the Denver Nuggets look like a team on the rise. But they're set to add a notable talent, as \error{Michael} \fph{Porter Jr.} has been medically cleared and according to Nick Kosmider of The Athletic he will \fph{play in the Summer League}...}\\
            
            \toprule
            \toprule
            
            \textbf{Gold Label}: {walmart announces next day delivery} & \textbf{Gold Label}: {new pokemon mobile game}\\
            \textbf{Human Supervision}: amazon orders \$day to buy amazon & \textbf{Human Supervision}: dena sword launch \\
            \textbf{Distant Supervision+Human Supervision}: walmart to offer next day delivery & \textbf{Distant Supervision+Human Supervision}: new pokemon mobile game \\
            \midrule
            
            \textbf{Article\textsubscript{1}}: \textbf{Walmart announces next-day delivery, firing back at Amazon}
            & \textbf{Article\textsubscript{1}}: \textbf{A New Pokemon Mobile Game Is In Development}\\
             {\fph{Walmart} is firing back. The biggest retailer in the world will now offer shoppers the option to have their online \fph{orders delivered the next day}, following \error{Amazon}... }
             &      {\error{Pokemon Sword and Shield} might have already been announced, but we now know there's another \fph{new Pokemon game} on the way from \error{DeNA}. The Japanese developer... } \\
                         \smallskip
             \textbf{Article\textsubscript{2}}: \textbf{Walmart Fires Back at Amazon, Announces Next-Day Delivery on 220,000 Items}
             \newline  {\fph{Walmart is rolling out free next-day delivery} on its most popular items, increasing the stakes in the retail shipping wars with \error{Amazon}. The nation's largest retailer said...} 
             &\smallskip \textbf{Article\textsubscript{2}}: \textbf{New Pokemon Mobile Game Announced!} \newline {Exciting news, Trainers! \error{DeNA}, a big player in the gaming world, has announced a \fph{new Pokemon mobile game} in their financial report! In partnership with The Pokemon Company, \error{DeNA} is working on a whole \fph{new Pokemon mobile game}...} \\
                         \smallskip
             \textbf{Article\textsubscript{3}}: \textbf{Walmart announces next-day delivery on 200K+ items in select markets}
             \newline              {This month, \error{Amazon} announced it's investing \$800 million in its warehouses and delivery infrastructure in order to double the speed of Prime shipping by reducing it to only one day. Now \fph{Walmart} is following suit with a \fph{one-day shipping announcement}...}
             & \smallskip \textbf{Article\textsubscript{3}}: \textbf{There's a new Pokemon mobile game in the works}
             \newline {A \fph{new Pokemon mobile game} is in the works. Japanese firm \error{DeNA} - behind Mario Kart Tour and Super Mario Run - is working on a brand new title based on the popular franchise with The Pokemon Company. According to TechRadar, \error{DeNa} said in an official document: [We] plan to launch a new and exciting smartphone game... }\\
            
            \bottomrule
        \end{tabular}
        }
\end{table*}
\subsection{Are attention modules robust to noises?}
\label{sec:exp_noise}
So far we have mostly considered models in ideal settings, where the news stories for training and testing are relatively clean and validated.
In real-world scenarios, however, automatically clustered news stories can be noisy, \ie they may include some irrelevant or even off-topic articles.
This is when the article-level attention module plays its role by assigning different weights to articles.

In this experiment, we intentionally add noises to the training and testing stories (both for \Newshead{} and \newsheaddist{}) by randomly replacing an article in each story with a randomly sampled article from the entire corpus.
Under the same architecture, we compare three different article-level attention designs: 
(1) Uniform Attention, which assigns equal weight to each article;
(2) Referee Attention, which determines article weights by an external ``referee'' vector, as introduced in Section~\ref{sec:article-attention} and Figure~\ref{fig:docatt};
(3) Our self-voting-based Attention.
We compare their performances under all pre-training settings.

Table~\ref{tb:ablation} shows the performance of different attention designs.
Among all attention modules, Referee Attention achieves the worst performance, which is as expected since Referee Attention can assign inappropriately large weights to noisy articles, severely corrupting the final output.
This would be further verified with real examples in Section \ref{sec:case}.
For comparison, the simple Uniform Attention module will at least avoid focusing on the wrong articles, and hence achieves better performance than Referee.
From the model perspective, the Referee model is more complex and is harder to train with limited resources.
This experiment also shows that, when on-topic articles dominate the story, even simple uniform attention can achieve satisfactory performance by integrating decoding outputs from different articles, but the traditional Referee Attention can produce risky results.
Our Self-voting-based Attention achieves best performances under all settings, thanks to its capability of leveraging the dynamic voting process between articles to emphasize shared common information and identify noises.
When tested without any pre-training (H), where the initial article representations are far from perfect, the performance gap between different attention designs becomes more vivid.

\section{Case Study}
\label{sec:case}
We conduct case studies to better understand the advantages of the proposed model.
Table~\ref{tb:case_clean} compares the full model with the variant without fine-tuning on human supervision and without distant supervision on the clean dataset.
As one may find, the model using pure distant supervision can already generate headlines of high quality in terms of representativeness and informativeness.
Fine-tuning on human labels further adjusts the length and style of the outputs by dropping undesirable verbosity such as full person names and prepositions.

The model without distant supervision fails to pay attention to important words, which results in less meaningful headlines that are often inarticulate.
For instance, in the news stories discussing Walmart's move against Amazon Prime delivery, the model accidentally generates a somewhat meaningless headline regarding Amazon, as the word ``Amazon'' appears frequently in the news stories.
In contrast, the full model generates a high-quality headline with very close semantics to the human label, replacing ``announces'' with ``to offer''.
A similar case can also be observed in the story concerning ``new Pokemon mobile game''.

Table \ref{tb:case_noisy} shows a representative comparison between Self-voting Attention and Referee Attention on the noisy dataset as described in Section~\ref{sec:exp_noise}.
When an outlier article is added to the story (article 2), the Referee Attention still assigned a relatively high attention weight to it, and hence introduced false information to the headline (``game seven'').
On the contrary, our model successfully identified the outlier through the dynamic voting process, and avoided adding noise to the generated headline.

During our study, we also find cases that can be improved.
Specifically, in some cases, headlines generated by existing models may focus on different information than human labels. 
For example, a story labeled as ``biden \emph{on china} tariffs'' by human experts is labeled by our algorithm as ``biden \emph{criticizes trump} on tariffs''.
Both are good summaries for the story, but focus on different aspects.
In the future, we may consider two directions to further satisfy personalized information needs.

\begin{table}[t]
    \footnotesize
        \caption{A case study on the noisy dataset.}
        \vspace{-.2cm}
        \label{tb:case_noisy}
        \scalebox{.9}{
        \begin{tabular}{p{0.92\columnwidth}}
            \toprule
            \textbf{Gold Label: {austin riley mlb debut}}\\
            \midrule
            \textbf{Referee Attention: \fph{austin riley} homers \error{in game 7}}  \\
            \textbf{Self-Voting Attention: \fph{austin riley mlb debut}} \\
            \midrule
            
             \textbf{Article\textsubscript{1}}: \textbf{Braves prospect Riley homers in 2nd MLB AB}\\
             {ATLANTA -- As \fph{Austin Riley} soaked in the excitement of highlighting his Major League \fph{debut} with a monstrous home run that helped the Braves claim a 4-0 win over the Cardinals on Wednesday night... }
             
             \medskip
             \textbf{Article\textsubscript{2}} (noise): \textbf{SMB completes PH Cup five-peat after gripping Game 7 win over Magnolia}\\
             {FIVE rings to adorn this San Miguel dynasty. The Beermen extended their reign in the PBA Philippine Cup in dramatic fashion, overcoming a 17-point deficit to beat Magnolia, 72-71, in a \error{Game Seven} to remember...} 
             
             \medskip
             \textbf{Article\textsubscript{3}}: \textbf{Called Up: Austin Riley}\\
             {Yesterday, the Braves called up \fph{Austin Riley}, who we ranked second in their system and 33rd in our Top 100. He continued his blazing hot 2019, going 1-for-3 in his big league \fph{debut} last night, including a home run...}\\
            \bottomrule
        \end{tabular}
        }
    \end{table}

\section{Related Work}
Three lines of research are closely related: document summarization, news headline generation, and language model pre-training.

\noindent\textbf{Single-Document Summarization (SDS)} has been studied for over six decades \cite{luhn1958automatic}.
Early extractive methods incorporate handcrafted features and graph-based structural information to select informative sentences to form the summary \cite{filatova2004event,mihalcea2005language,woodsend2010automatic}.
Neural extractive models achieve significant improvement by taking advantage of effective representation learning \cite{cheng2016neural,nallapati2017summarunner,narayan2018ranking,zhang-etal-2018-neural}.
Recent success of seq2seq models has inspired various abstractive summarization methods with an end-to-end encoder-decoder architecture to achieve new state-of-the-art performance \cite{rush-etal-2015-neural}.
The encoder module represents input sentences with word embeddings, linguistic features \cite{nallapati2016abstractive}, and abstract meaning representation \cite{takase2016neural,liao2018abstract}. 
The input sequence is encoded to an intermediate representation and decoded to target sequence with an RNN or Transformer and their variants \cite{nallapati2016abstractive,Chen:2016:DNN:3060832.3061006,li2017deep,liu2018generating}.
To alleviate the Out-Of-Vocabulary (OOV) issue, prior works also incorporate various copy mechanisms \cite{vinyals2015pointer,gu2016incorporating,see2017get} in the framework.
 Recent works improve the quality of generated summaries in length \cite{kikuchi2016controlling,makino2019global,narayan2018don} and informativeness \cite{pasunuru2018multi,li2018improving,you2019improving,peyrard2019simple}.
 In addition, many alternative evaluation metrics~\cite{mao2019facet,narayan2019highres,xenouleas-etal-2019-sum} have been proposed recently due to the limits of ROUGE.

\noindent\textbf{Multi-Document Summarization (MDS)}, on the other hand, aims to generate summaries for a set of documents.
Early works on MDS explore both extractive \cite{carbonell1998use,radev2004centroid,erkan2004lexrank,mihalcea2004textrank,haghighi2009exploring} and abstractive methods \cite{mckeown1995generating,radev1998generating,barzilay1999information,ganesan2010opinosis}.
End-to-end abstractive models for MDS are limited by a lack of large-scale annotated datasets.
Recent works either try to leverage external resources \cite{liu2018generating,liu-lapata-2019-hierarchical} or adapt single-document summarization models to MDS tasks \cite{baumel2018query,zhang2018adapting,lebanoff2018adapting}.
The recently developed multi-news dataset \cite{fabbri-etal-2019-multi} provides the first large-scale training data for supervised MDS, while such a dataset is still absent for the multi-document headline generation task. 

\noindent\textbf{Headline Generation} is a special task of Document Summarization to generate headline-style abstracts of articles \cite{shen2017recent}, which are usually shorter than a sentence \cite{banko2000headline}.
Over the past decade, both rule-based \cite{dorr2003hedge}, compression-based \cite{filippova2013overcoming,filippova2015sentence} and statistical-based methods \cite{banko2000headline,zajic2002automatic} have been explored with handcrafted features and linguistic rules.
Recent state-of-the-art headline generation models are dominated by end-to-end encoder-decoder architectures \cite{rush-etal-2015-neural,hayashi2018headline,zhang2018question,gavrilov2019self,murao2019case}.
Similar to summarization models, the encoder module considers different formats of input representation including word position embedding \cite{chopra2016abstractive}, abstractive meaning representations \cite{takase2016neural} and other linguistic features \cite{nallapati2016abstractive}.
Pointer networks \cite{nallapati2016abstractive} and length-controlling mechanisms \cite{kikuchi2016controlling} are also developed for this task.
However, to the best of our knowledge, the task of headline generation for multiple documents has barely been explored before.

\noindent\textbf{Language Model Pre-training} has been proven to be effective in boosting the performance of various NLP tasks with little cost \cite{radfordimproving,peters2018deep,devlin2019bert,zhang-etal-2019-ernie,sun2019ernie,radford2019language}.
Recent works applying pre-trained language models also achieve significant success in summarization and headline generation tasks \cite{song2019mass,rothe2019leveraging,zhang2019pretraining}.
In this work, we study how pre-training at different levels can benefit multi-document headline generation.
\section{Conclusions}
In this work, we propose to study the problem of generating headline-style abstracts of articles in the context of news stories.
For standard research and evaluation, we publish the first benchmark dataset, \newshead{}, curated by human experts for this task.
The slow and expensive curation process, however, calls for effort-light solutions to achieve abundant training data from web-scale unlabeled corpus.
To this end, we propose to automatically annotate unseen news stories by distant supervision where a representative article title is selected as the story headline.
Together with a multi-level pre-training framework, this new data augmentation approach provides us with a $6$ times larger dataset without human curator and enables us to fully leverage the power of Transformer-based model. A novel self-voting-based article attention is applied afterward to better extract salient information shared by multiple articles. Extensive experiments have been conducted to verify \nhnet{}'s superior performance and its robustness to potential noises in news stories.

\bibliographystyle{ACM-Reference-Format}
\bibliography{acmart}


\end{document}